\documentclass[sigplan]{acmart}
\AtBeginDocument{%
  }

\renewcommand\footnotetextcopyrightpermission[1]{} 
\setcopyright{none}
\copyrightyear{}
\acmYear{}
\acmDOI{}
\acmConference[]{}{}{}
\acmISBN{}

\begin{document}

\title{Tutoring Large Language Models to be Domain-adaptive, Precise, and Safe}


\author{Somnath Banerjee}
\affiliation{%
  \institution{Indian Institute of Technology Kharagpur}
  \country{India}
}

\author{Animesh Mukherjee}
\authornote{Supervisor}
\affiliation{%
  \institution{Indian Institute of Technology Kharagpur}
  \country{India}
}









\begin{abstract}
The overarching research direction of this work is the development of a ``\textit{Responsible Intelligence}'' framework designed to reconcile the immense generative power of Large Language Models (LLMs) with the stringent requirements of real-world deployment. As these models become a transformative force in artificial intelligence, there is an urgent need to move beyond general-purpose architectures toward systems that are contextually aware, inherently safer, and deeply respectful of global cultural nuances. This research navigates three interconnected threads: domain adaptation to ensure technical precision, ethical rigor to mitigate adversarial vulnerabilities, and cultural/multilingual alignment to promote global inclusivity. The methodological trajectory moves from classical supervised adaptation for task-specific demands to decoding-time alignment for safety, finally leveraging human feedback and preference modeling to achieve sociolinguistic acuity .

\end{abstract}

\maketitle

\section{Motivation}

Despite the fluency and coherence of modern LLMs, their transition from laboratory demos to critical infrastructure has unveiled systemic risks. General-purpose models frequently lack the fine-grained expertise required for high-stakes domains like software engineering, where a failure to understand intricate naming conventions can lead to costly errors in production environments . Furthermore, the instruction-following design of these models introduces a dangerous tension between expressive utility and control; static, ``one-size-fits-all'' guardrails are often bypassed by sophisticated adversarial strategies, such as "jailbreaking" through structured pseudocode . Finally, the predominantly Western-centric training data of current LLMs results in a cultural mismatch, where models inadvertently propagate stereotypes or misinterpret the social norms and taboos of diverse global communities . If left unaddressed, these challenges threaten the fundamental trust and reliability required for human-centric AI. \\I argue that LLMs can be made reliable for web-facing deployment by tutoring them with (i) domain grounding for technical precision, (ii) robust safety steering against structured jailbreaks, and (iii) culturally-aware multilingual alignment. My dissertation will unify these into a single framework and evaluate it on technical QA, adversarial safety, and cultural-harm benchmarks.\\
\noindent \textbf{Definition (Tutoring)}: In this paper, ``tutoring'' means targeted guidance that improves reliability without full retraining—via (a) data/label tutoring (weak supervision/active learning), (b) context tutoring (retrieval/graph grounding), and (c) behavior tutoring (decoding-time or parameter steering).

\section{Problem Statement}
Since the beginning of my PhD, I have addressed the following research questions--\\
\noindent \textbf{RQ1: How can LLMs be tutored to achieve domain-adaptive precision in specialized, low-resource technical ecosystems?}\\
The fundamental challenge within this research question is the lack of "technical literacy" in general-purpose LLMs. In specialized fields like software engineering, precision is not just a preference but a functional requirement. General models often struggle with the unique linguistic characteristics of technical documentation—such as specific versioning syntax, repository naming conventions, and complex dependency relationships. Without a mechanism to ground these models in domain-specific entities and structured knowledge, they remain prone to "hallucinations," producing code or advice that appears superficially correct but is functionally invalid or obsolete.\\ 
\noindent \textbf{RQ2: How can model guardrails be reinforced to mitigate safety vulnerabilities without stifling expressive capabilities?}\\
The fundamental challenge in RQ2 is the inherent tension between an LLM's ability to follow complex, open-ended instructions and the necessity of preventing the generation of harmful or unethical content. While safety training (such as RLHF) has improved model behavior, these systems remain susceptible to "jailbreaking" techniques where malicious users circumvent guardrails by masking harmful intent within structured, instruction-centric formats like pseudocode . The problem is that current safety mechanisms are often optimized for natural language and fail to recognize unethical patterns when they are abstracted into symbolic or logical structures.\\
\noindent \textbf{RQ3: What mechanisms can ensure cultural and multilingual alignment for globally diverse and inclusive AI interactions?}\\
The core problem in RQ3 is the "cultural mismatch" that occurs when LLMs, predominantly trained on Anglophone or Western-centric data, are deployed globally. These models often fail to respect the diverse social norms, taboos, and etiquette of non-Western communities, leading to "cultural harm" through the perpetuation of stereotypes or the misinterpretation of region-specific idioms . Furthermore, safety guardrails that work effectively in English often degrade in low-resource or code-switching multilingual environments, creating an uneven safety landscape for global users.

\section{State of the Art}
\subsection{Domain adaptation and context precision}

\noindent \textbf{Part I: Data Scarcity and Naming Complexity in Software NER} Traditional Named Entity Recognition (NER) models are typically trained on generic datasets (like news or Wikipedia), which fail to capture the nuances of software-related entities. There is a severe scarcity of "gold-labeled" data for software-specific tasks, and the cost of manual annotation by experts is prohibitively high. Furthermore, existing models struggle with the high degree of ambiguity in technical naming; for instance, a model might misidentify a software package name as a common noun or fail to recognize the structural importance of a specific error code, limiting its ability to assist in automated bug-fixing or documentation.\\
\noindent \textbf{Part II: Limitations of Flat-Text Retrieval in Technical Q\&A} Current Retrieval-Augmented Generation (RAG) systems primarily rely on flat-text similarity, which is insufficient for answering complex technical questions on platforms like AskUbuntu or ServerFault. These systems often fail to capture the "multi-hop" relationships between different technical components—such as how a specific OS version interacts with a particular library. Because standard retrieval mechanisms lack a structured understanding of these dependencies, the LLM is often forced to fill in the gaps with plausible-sounding but factually incorrect information, leading to unreliable technical support.

\subsection{Safety and ethical considerations}
\noindent \textbf{Part I: Lack of Robust Benchmarks for Symbolic Logic Attacks:} Traditional safety evaluations primarily focus on narrative text and direct harmful queries. There is a critical absence of dedicated benchmark datasets to test model robustness against "instruction-centric" responses, where a model might refuse a harmful request in plain text but fulfill it if the request is formatted as a pseudocode logic problem. This leaves a massive security gap where a model's advanced reasoning capabilities are turned against its own safety alignment.\\
\noindent \textbf{Part II: Limitations of Static and Post-Hoc Guardrails:} Most industrial safety solutions rely on static content-filtering mechanisms that block specific keywords or topics. These "one-size-fits-all" filters are context-blind and often fail to block sophisticated adversarial prompts or, conversely, over-block innocuous information . There is a need for a dynamic safety mechanism that can adjust the model's generation process in real-time based on the evolving context of the conversation rather than applying a crude filter after the fact.

\subsection{Cultural and multilingual alignment}
\noindent \textbf{Part I: Subjective and Scanty Cultural Harm Datasets:} Existing safety benchmarks rarely account for the subtle, context-dependent nature of cultural harm, which varies drastically across geographic regions. There is a lack of structured datasets that map specific "contextual factors"—such as religious sensitivities or local historical taboos—to model evaluation. Without these, models cannot be effectively fine-tuned to recognize and avoid outputs that might be socially acceptable in one culture but deeply offensive in another.\\
\noindent \textbf{Part II: Inefficiency of Global-Only Parameter Steering:} Most multilingual safety efforts rely on "bridging strategies," like translating all non-English queries into English before applying filters. These methods are computationally expensive and often distort the original meaning. There is a technical gap in identifying and modifying the specific internal components of a model (e.g., individual attention heads) that are responsible for language-specific harmful behaviors without requiring a full retraining of the model for every new language.

\section{Proposed Approach}
\subsection{Domain adaptation and context precision}
\noindent \textbf{Part I: The \textsc{DistALANER}} Framework To address the data scarcity and naming challenges, this research introduces \textsc{DistALANER}, a framework that synthesizes distant supervision and active ~\cite{10.1007/978-3-031-70381-2_20}. By leveraging weakly supervised data from 170,000 bug reports and software manuals, the system bootstraps the model to recognize nine critical software-specific entity types (e.g., Application, Library, OS, and Architecture). This approach allows for the high-precision extraction of technical entities with minimal human intervention, demonstrating that LLMs can be effectively "tutored" on specialized technical vocabularies even when high-quality labeled data is unavailable.\\
\noindent \textbf{Part II: The \textsc{GraphContextGen}} Framework To solve the problem of factual grounding in technical generation, the thesis proposes \textsc{GraphContextGen}~\cite{banerjee-etal-2024-context}. This system moves beyond simple text retrieval by integrating graph-based retrieval mechanisms—specifically using Personalized PageRank (PPR)—over structured knowledge from Wikidata. By representing technical entities and their relationships as a graph, the framework infuses the LLM with a "factually robust" context. This ensures that the generated responses are anchored in structural reality, significantly reducing hallucinations and providing contextually coherent solutions for technical users on community Q\&A platforms.

\subsection{Safety and ethical considerations}
\noindent \textbf{Part I: The \textsc{TechHazaraQA} Benchmark:} This research contributes \textsc{TechHazaraQA}, a first-of-its-kind benchmark dataset containing approximately 7,745 sensitive and unethical queries across seven high-risk technological areas, including biotechnology, cybersecurity, and nuclear technology~\cite{Banerjee_Layek_Hazra_Mukherjee_2025}. This dataset allows researchers to systematically audit LLMs by presenting queries that can be answered in both plain text and pseudocode, revealing that structured prompts can increase unethical generations by 2-38\% even in extensively safety-trained models.\\
\noindent \textbf{Part II: The SafeInfer Alignment Framework:} To provide a more robust defense, the thesis introduces SafeInfer, a context-adaptive, decoding-time safety alignment framework~\cite{Banerjee_Layek_Tripathy_Kumar_Mukherjee_Hazra_2025}. SafeInfer operates in two phases: it first steers the model's internal latent space away from harmful representations using "safety amplification" vectors, and then applies a "Safety Guided Decoding Strategy" (sGDS) that dynamically adjusts token probabilities to favor safe outputs . This dual approach significantly reduces the Attack Success Rate (ASR) to as low as 1.09\% on standard benchmarks while preserving the model's general utility and expressive fluency.

\subsection{Cultural and multilingual alignment}

\noindent \textbf{Part I: The Cultural Kaleidoscope Datasets and Alignment:} This research provides two major resources: a Cultural Harm Evaluation Dataset (~15,000 queries) and a Culturally Aligned Preference Dataset~\cite{banerjee-etal-2025-navigating}. These datasets cover 11 major global cultures and 12 sensitive topics, using "crescendo-based" multi-turn conversations to test for hidden biases. By applying Offline Reward-based Preference Optimization (ORPO), the thesis demonstrates that models can be fine-tuned to reduce harmful outputs dramatically—for instance, dropping Mistral-7B's cultural harm rate from 71.96\% to 3.07\%.\\
\noindent \textbf{Part II: The Soteria Parameter Steering Framework:} To address multilingual safety efficiently, the thesis introduces Soteria, a lightweight method that identifies language-specific "functional heads" within the model's architecture~\cite{banerjee-etal-2025-soteria}. By selectively tuning only ~3\% of the model’s parameters to "subtract" harm directions identified in specific languages, Soteria reduces policy violations across a broad linguistic spectrum—from high-resource languages like Chinese to low-resource ones like Bengali (ASR reduced from 0.46 to 0.29)—all while preserving the model’s overall performance on general benchmarks.

\section{Conclusion \& Future Work}

\subsection{A Unified Vision: Responsible Intelligence}
This dissertation moves beyond fragmented AI safety by proposing a \textbf{Responsible Intelligence} framework. The core vision is that technical precision, ethical robustness, and cultural inclusivity are not independent goals but are deeply interconnected. By ``tutoring'' models, using targeted, resource-efficient interventions like decoding-time steering and lightweight parameter updates~\cite{Banerjee_Layek_Tripathy_Kumar_Mukherjee_Hazra_2025, hazra-etal-2024-safety}, we can reconcile the generative power of LLMs with the stringent reliability required for critical web infrastructure. This trajectory shifts the paradigm from rigid, post-hoc content filtering toward a more fluid, context-aware intelligence that masters the nuances of symbolic logic and technical ecosystems.

\subsection{Significance and Potential Impact}
The significance of this work lies in its multi-layered impact on the deployment of LLMs:
\begin{itemize}
    \item \textbf{Industrial Technical Precision:} By grounding models in structured knowledge (e.g., \textsc{GraphContextGen}), this research provides a blueprint for deploying LLMs in high-stakes domains like software engineering, where ``hallucinated'' advice can lead to costly production errors~\cite{banerjee-etal-2024-context, 10.1007/978-3-031-70381-2_20}.
    \item \textbf{Ethical Robustness against Symbolic Attacks:} The introduction of \textsc{TechHazardQA} and \textsc{SafeInfer} closes a massive security gap by defending against instruction-centric and pseudocode-based jailbreaks that currently bypass standard natural language guardrails~\cite{Banerjee_Layek_Hazra_Mukherjee_2025, Banerjee_Layek_Tripathy_Kumar_Mukherjee_Hazra_2025}.
    \item \textbf{Global Inclusivity and Safety Equity:} Through \textsc{Soteria} and the \textsc{Cultural Kaleidoscope} datasets, this work ensures that safety guardrails are not ``Western-centric'' but are instead deeply respectful of the social norms and taboos of diverse global communities~\cite{banerjee-etal-2025-navigating, banerjee-etal-2025-soteria}.
    \item \textbf{Resource-Efficient Sustainability:} Our focus on ``tutoring'' mechanisms allows for continuous model alignment without the prohibitive computational and environmental costs of full-scale retraining~\cite{banerjee-etal-2025-navigating,banerjee-etal-2025-soteria, hazra-etal-2024-safety,Banerjee_Layek_Tripathy_Kumar_Mukherjee_Hazra_2025}.
\end{itemize}

\subsection{Current Focus \& Future Directions}
The current phase focuses on consolidating these capabilities into a unified, resource-efficient architecture for industrial deployment. Future research will explore \textbf{mechanistic interpretability} to ensure that the internal ``neurons'' of future AI systems remain fundamentally aligned with human values and technical facts across every global community. I am also investigating the ripple effects of model editing~\cite{banerjee-etal-2025-breaking} on cross-linguistic performance to ensure that localized updates do not compromise global reliability.

\bibliographystyle{ACM-Reference-Format}
\bibliography{sample-base}


\end{document}